\documentclass[10pt,twocolumn,letterpaper]{article}

\usepackage{cvpr}
\usepackage{times}
\usepackage{epsfig}
\usepackage{graphicx}
\usepackage{amsmath}
\usepackage{amssymb}
\usepackage{multirow}
\usepackage{wasysym} 
\usepackage[normalem]{ulem}
\usepackage{subcaption}
\usepackage{ctable}

\newcommand{\head}[1]{{\noindent\textbf{#1}}}

\newcommand{\ours}{Discriminative Motion Cue Network\xspace}
\newcommand{\oursnickname}{DMC-Net\xspace}

\newcommand{\vx}{\mathbf{x}}

\makeatletter
\DeclareRobustCommand\onedot{\futurelet\@let@token\@onedot}
\def\@onedot{\ifx\@let@token.\else.\null\fi\xspace}
\def\eg{e.g\onedot} 
\def\ie{\emph{i.e}\onedot}

\def\etal{\emph{et~al}\onedot}

\makeatother

\usepackage[pagebackref=true,breaklinks=true,letterpaper=true,colorlinks,bookmarks=false]{hyperref}

\cvprfinalcopy 

\def\cvprPaperID{1054} 

\ifcvprfinal\fi

\begin{document}

%
\title{DMC-Net: Generating Discriminative Motion Cues for \\ Fast Compressed Video Action Recognition}
\author{Zheng Shou$^{1,2}$
\and
Xudong Lin$^2$
\and
Yannis Kalantidis$^1$
\and 
Laura Sevilla-Lara$^{1,3}$
\and
Marcus Rohrbach$^1$
\and
Shih-Fu Chang$^2$
\and
Zhicheng Yan$^1$
\and
\\
$^{1}$Facebook AI \hspace{30pt} $^{2}$Columbia University \hspace{30pt} $^{3}$Univesity of Edinburgh 
}

\maketitle
\thispagestyle{empty}

\let\thefootnote\relax\footnotetext{This work was partially done when Zheng Shou interned at Facebook.}

\begin{abstract}

Motion has shown to be useful for video understanding, where motion is typically represented by optical flow. However, computing flow from video frames is very time-consuming.
Recent works directly leverage the motion vectors and residuals readily available in the compressed video to represent motion at no cost.
While this avoids flow computation, it also hurts accuracy since the motion vector is noisy and has substantially reduced resolution, which makes it a less discriminative motion representation. 
To remedy these issues, we propose a lightweight generator network, which reduces noises in motion vectors and captures fine motion details, achieving a more Discriminative Motion Cue \textbf{(DMC)} representation.
Since optical flow is a more accurate motion representation, we train the DMC generator to approximate flow using a reconstruction loss and an adversarial loss, jointly with the downstream action classification task.
Extensive evaluations on three action recognition benchmarks (HMDB-51, UCF-101, and a subset of Kinetics) confirm the effectiveness of our method.
Our full system, consisting of the generator and the classifier, is coined as \textbf{\oursnickname} which obtains high accuracy close to that of using flow and runs two orders of magnitude faster than using optical flow at inference time.

\end{abstract}


\section{Introduction}
\label{sec:introduction}

\begin{figure}[h]
\centering
\centerline{\includegraphics[width=0.95\linewidth]{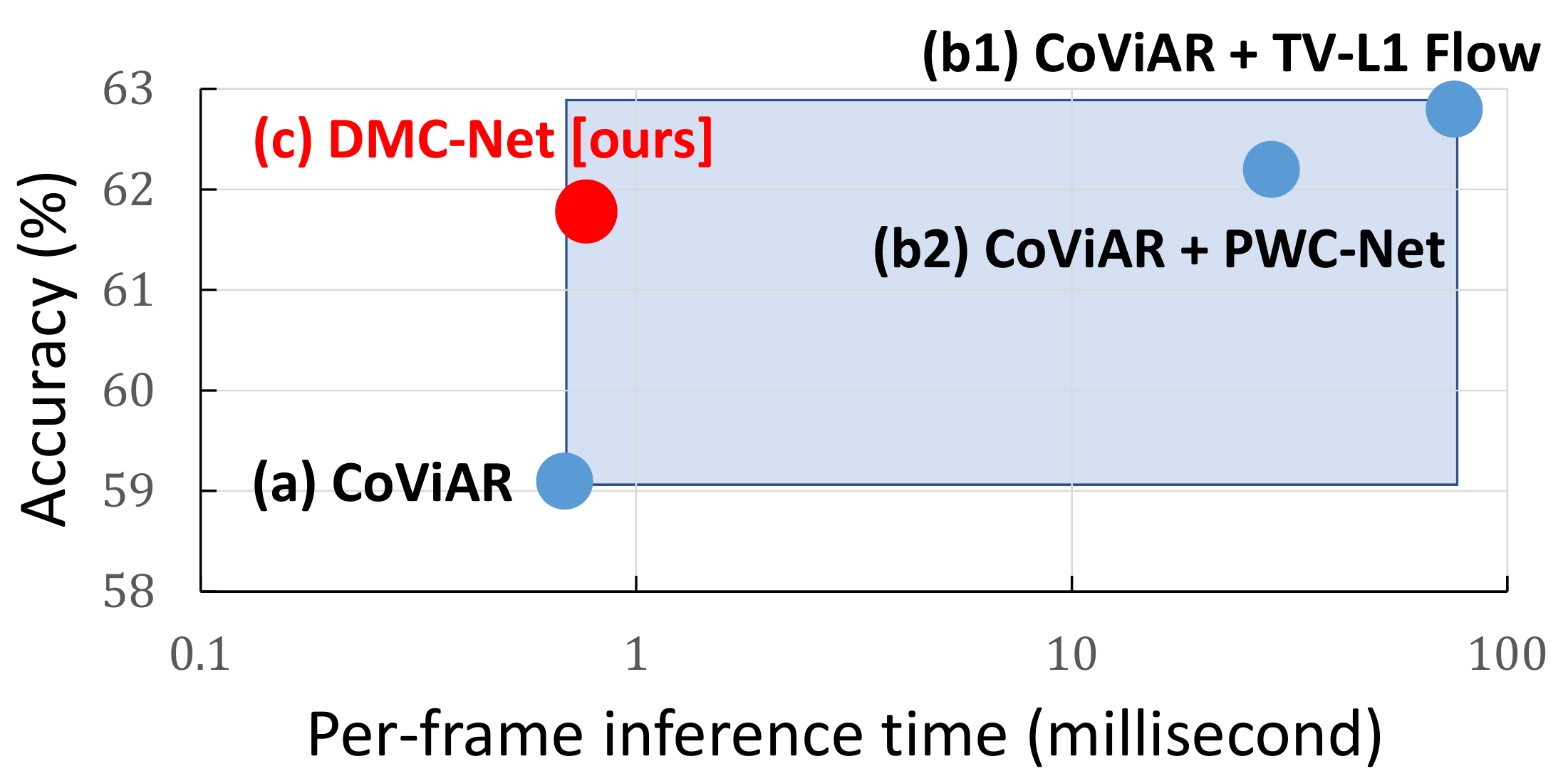}}
   \caption{Comparing inference time and accuracy for different methods on HMDB-51. \textbf{(a)} Compressed video based method \textit{CoViAR}~\cite{wu2018coviar} is very fast.
   \textbf{(b)} But in order to reach high accuracy, \textit{CoViAR} has to follow two-stream networks to add the costly optical flow computation, either using TV-L1~\cite{TVL1} or PWC-Net~\cite{Sun2018PWC-Net}.
   \textbf{(c)} The proposed \textit{DMC-Net} not only operates exclusively in the compressed domain, but also is able to achieve high accuracy while being two orders of magnitude faster than methods that use optical flow. The blue box denotes the improvement room from \textit{CoViAR} to \textit{CoViAR + TV-L1 Flow}; x-axis is in logarithmic scale. }
   
   \label{fig:improvement space}
\end{figure}

\begin{figure}[t!]
\vspace{-0.5em}
\centering
\begin{subfigure}{.9\linewidth}
  \centering
  \includegraphics[width=1.0\linewidth]{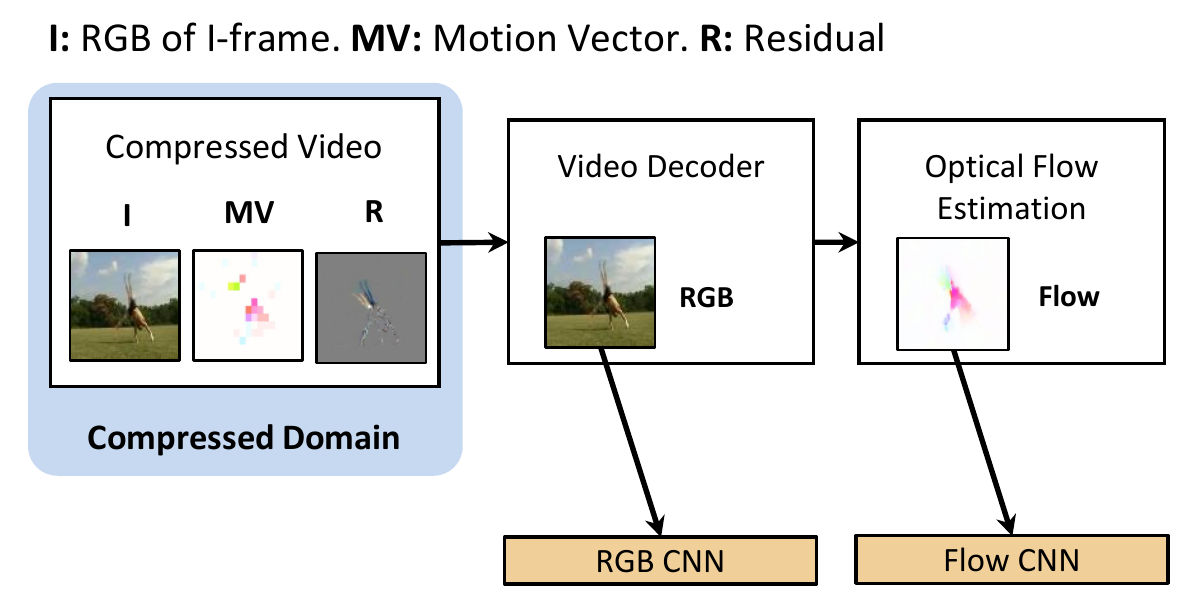}
  \caption{Two-stream network (TSN)~\cite{Simonyan14b}}
  \label{fig:sub1}
\end{subfigure}%
\hspace{5pt}
\begin{subfigure}{.9\linewidth}
  \centering
  \includegraphics[width=1.0\linewidth]{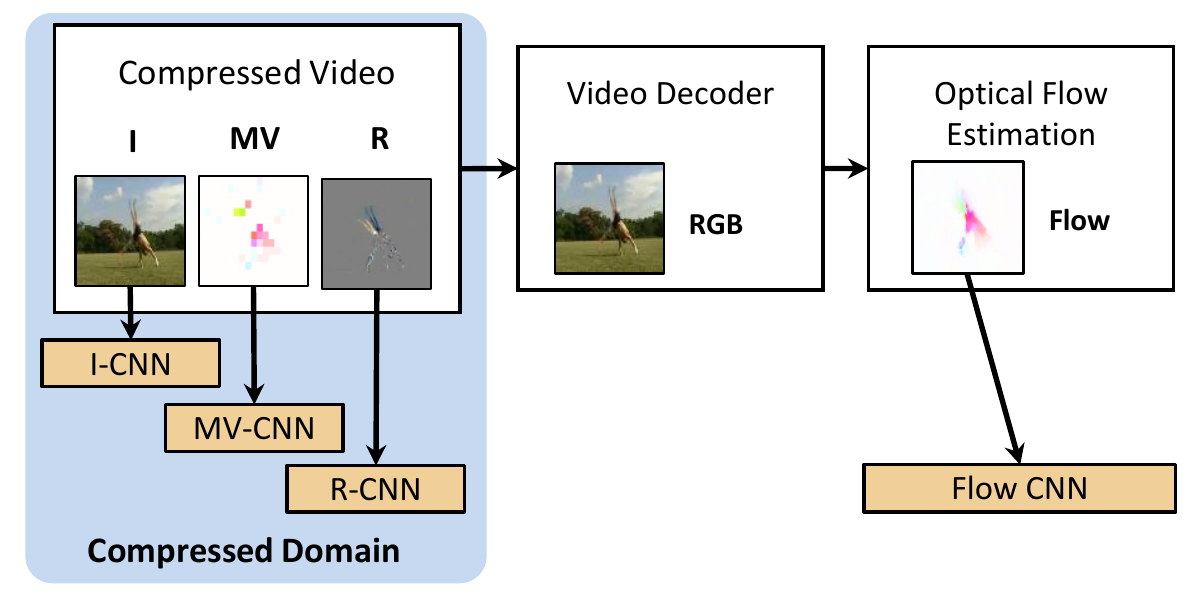}
  \caption{CoViAR~\cite{wu2018coviar} + Flow}
  \label{fig:sub2}
\end{subfigure}
\hspace{5pt}
\begin{subfigure}{.9\linewidth}
  \centering
  \includegraphics[width=1.0\linewidth]{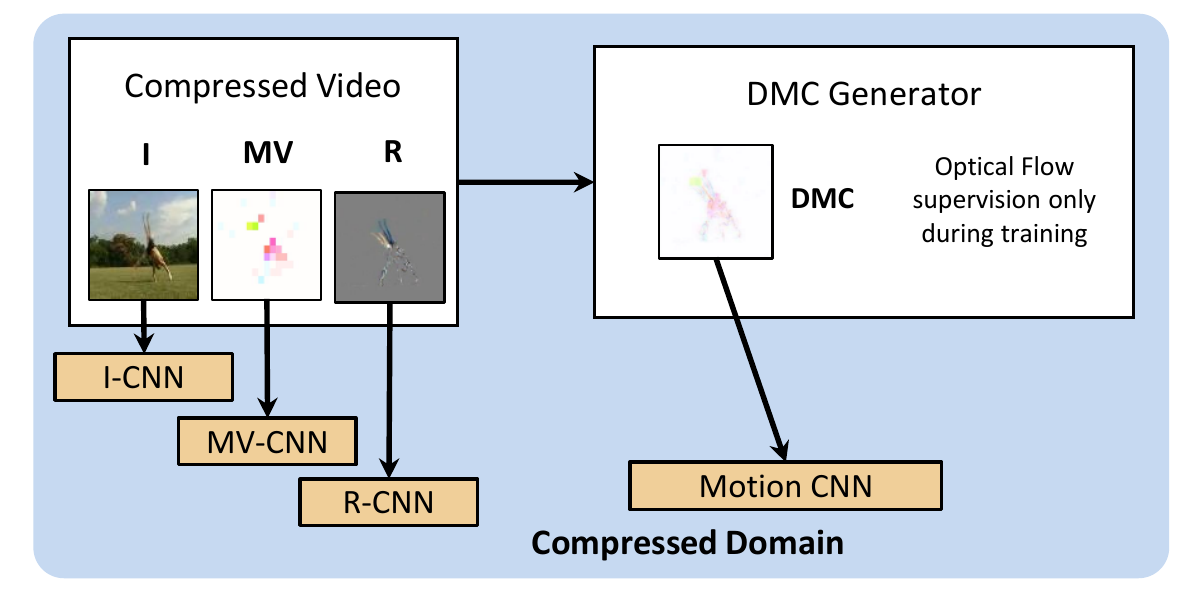}
  \caption{DMC-Net (ours)}
  \label{fig:sub3}
\end{subfigure}
\caption{Illustrations of (a) the two-stream network~\cite{Simonyan14b}, (b) the recent CoViAR~\cite{wu2018coviar} method that achieves high accuracy via fusing compressed video data and optical flow, and (c) our proposed DMC-Net. Unlike \textit{CoViAR+Flow} that requires video decoding of RGB images and flow estimation, our DMC-Net operates exclusively in the compressed domain at inference time while using optical flow to learn to capture discriminative motion cues at training time.}
\label{fig:intro_methods}
\end{figure}


Video is a rich source of visual content as it not only contains appearance information in individual frames, but also temporal motion information across consecutive frames. Previous work has shown that modeling motion is important to various video analysis tasks, such as action recognition~\cite{Simonyan14b, wang2013action, laptev2008learning}, action localization~\cite{shou2018autoloc,shou2017cdc,scnn_shou_wang_chang_cvpr16,chao2018rethinking,shou2018online,lin2017single,lin2017temporal} and video summarization~\cite{tejero2018summarization, mendi2013sports}.
Currently, methods achieving state-of-the-art results usually follow the two-stream network framework~\cite{Simonyan14b, Kinetics, tran2018closer},
which consists of two Convolutional Neural Networks (CNNs), one for the decoded RGB images and one for optical flow, as shown in Figure \ref{fig:sub1}. These networks can operate on either single frames (2D inputs) or clips (3D inputs) and may utilize 3D spatiotemporal convolutions~\cite{tran2015learning, tran2018closer}.


Extracting optical flow, however, is very slow and often dominates the overall processing time of video analysis tasks. 
Recent work~\cite{wu2018coviar,zhang2018real_DTMV,ZhangWWQW16_MVCNN} avoids optical flow computation by exploiting the motion information from compressed videos encoded by standards like MPEG-4~\cite{le1991mpeg}. 
Such methods utilize the motion vectors and residuals already present in the compressed video to model motion. The recently proposed CoViAR~\cite{wu2018coviar} method, for example, contains three independent CNNs operating over three modalities in the compressed video, \ie RGB image of I-frame (\textbf{I}), low-resolution Motion Vector (\textbf{MV}) and Residual (\textbf{R}). The predictions from individual CNNs are combined by late fusion. CoViAR runs extremely fast while modeling motion features (see Figure~\ref{fig:sub2}). 
However, in order to achieve state-of-the-art accuracy, late fusion with optical flow is further needed (see Figure \ref{fig:improvement space}). 

This performance gap is due to the motion vector being less informative and discriminative than flow.
First, the spatial resolution of the motion vector is substantially reduced (\ie 16x) during video encoding, and fine motion details, which are important to discriminate actions, are permanently lost. Second, employing two CNNs to process motion vectors and residuals separately ignores the strong interaction between them.
Because the residual is computed as the difference between the raw RGB image and its reference frame warped by the motion vector.
The residual is often well-aligned with the boundary of moving object, which is more important than the motion at other locations for action recognition according to \cite{sevilla2017integration}.
Jointly modeling motion vectors and residuals, which can be viewed as coarse-scale and fine-scale motion feature respectively, can exploit the encoded motion information more effectively. 

To address those issues, we propose a novel approach to learn to 
generate a \textbf{Discriminative Motion Cue (DMC)} representation by refining the noisy and coarse motion vectors.
We develop a lightweight DMC generator network that operates on stacked motion vectors and residuals.
This generator requires training signals from different sources to capture discriminative motion cues and incorporate high-level recognition knowledge.
In particular, since flow contains high resolution and accurate motion information, we encourage the generated DMC to resemble optical flow by a pixel-level reconstruction loss. We also use an adversarial loss~\cite{goodfellow2014generative} to approximate the distribution of optical flow.
Finally, the DMC generator is also supervised by the downstream action recognition classifier in an end-to-end manner, allowing it to learn motion cues that are discriminative for recognition.

During inference, the DMC generator is extremely efficient with merely \emph{0.23} GFLOPs, and takes only \emph{0.106 ms} per frame which is negligible compared with the time cost of using flow.
In Figure~\ref{fig:sub3}, we call our full model \emph{DMC-Net}.
Although optical flow is required during training, our method operates \textit{exclusively in the compressed domain} at inference time and runs two orders of magnitude faster than methods using optical flow,
as shown in Figure~\ref{fig:improvement space}. Our contributions are summarized as follows: 

\begin{itemize}
    \setlength{\itemsep}{-0.2\baselineskip}
    \item We propose \oursnickname, a novel and highly efficient framework that operates exclusively in the compressed video domain and is able to achieve high accuracy without requiring optical flow estimation.
    \item We design a lightweight generator network that can learn to predict discriminative motion cues by using optical flow as supervision and being trained jointly with action classifier. 
    During inference, it runs two orders of magnitude faster than estimating flow.
    \item We extensively evaluate \oursnickname on 3 action recognition benchmarks, namely HMDB-51~\cite{kuehne2011hmdb}, UCF-101~\cite{soomro2012ucf101} and a subset of Kinetics~\cite{kay2017kinetics}, and demonstrate that it can significantly shorten the performance gap between state-of-the-art compressed video based methods with and without optical flow.
\end{itemize}

\section{Related Work}
\label{sec:related}

\noindent \textbf{Video Action Recognition.}
Advances in action recognition are largely driven by the success of 2D ConvNets in image recognition. The original Two-Stream Network~\cite{Simonyan14b} employs separate 2D ConvNets to process RGB frames and optical flow, and merges their predictions by late fusion. Distinct from image, video possesses temporal structure and motion information which are important for video analysis. This motivates researchers to model them more effectively, such as 3D ConvNets~\cite{tran2015learning, Kinetics}, Temporal Segment Network (TSN)~\cite{TSN}, dynamic image networks~\cite{bilen2018action}, and Non-Local Network~\cite{wang2017non}.
Despite the enormous amount of effort on modeling motion via temporal convolution, 3D ConvNets can still achieve higher accuracy when fused with optical flow~\cite{Kinetics, tran2018closer}, which is unfortunately expensive to compute.

\noindent \textbf{Compressed Video Action Recognition.}
Recently, a number of approaches that utilize the information present in the compressed video domain have been proposed. In the pioneering works~\cite{ZhangWWQW16_MVCNN,zhang2018real_DTMV}, Zhang \etal replace the optical flow stream in two-stream methods by a motion vector stream, but it still needed to decode RGB image for P-frame and ignored other motion-encoding modalities in compressed videos such as the residual maps.
More recently, the CoViAR method~\cite{wu2018coviar} proposed to exploit all data modalities in compressed videos, \ie RGB I-frames, motion vectors and residuals to bypass RGB frame decoding. However, CoViAR fails to achieve performance comparable to that of two-stream methods, mainly due to the low-resolution of the motion vectors and the fact that motion vectors and residuals, although highly related, are processed by independent networks.
We argue that, when properly exploited, the compressed video modalities have enough signal to allow us to capture more discriminative motion representation. We therefore explicitly learn such representation as opposed to relying on optical flow during inference.  

\noindent \textbf{Motion Representation and Optical Flow Estimation.}
Traditional optical flow estimation methods explicitly model the displacement at each pixel between successive frames~\cite{horn1981determining, zach2007duality, 2008learningflow, bruhn2005lucas}. In the last years CNNs have successfully been trained to estimate the optical flow, including FlowNet~\cite{dosovitskiy2015flownet, ilg2017flownet}, SpyNet~\cite{ranjan2017optical} and PWC-Net~\cite{Sun2018PWC-Net}, 
and achieve low End-Point Error (EPE) on challenging benchmarks, such as MPI Sintel~\cite{Butler:ECCV:2012} and KITTI 2015~\cite{Menze2015ISA}. Im2Flow work \cite{gaoim2flow} also shows optical flow can be hallucinated from still images. Recent work however, shows that accuracy of optical flow does not strongly correlate with accuracy of video recognition~\cite{gcpr18}. Thus, motion representation learning methods focus more on generating discriminative motion cues. Fan \etal~\cite{fan2018end} proposed to transform TV-L1 optical flow algorithm into a trainable sub-network, which can be jointly trained with downstream recognition network. Ng \etal~\cite{ng2018actionflownet} employs fully convolutional ResNet model to generate pixel-wise prediction of optical flow, and can be jointly trained with recognition network.
Unlike optical flow estimation methods, our method does not aim to reduce EPE error. Also different from 
all above methods of motion representation learning which take decoded RGB frames as input, our method refines motion vectors in the compressed domain, and requires much less model capacity to generate discriminative motion cues.

\section{Approach}
\label{sec:model}
\begin{figure*}[t!]
\vspace{-4em}
\begin{center}
\includegraphics[width=0.9\linewidth]{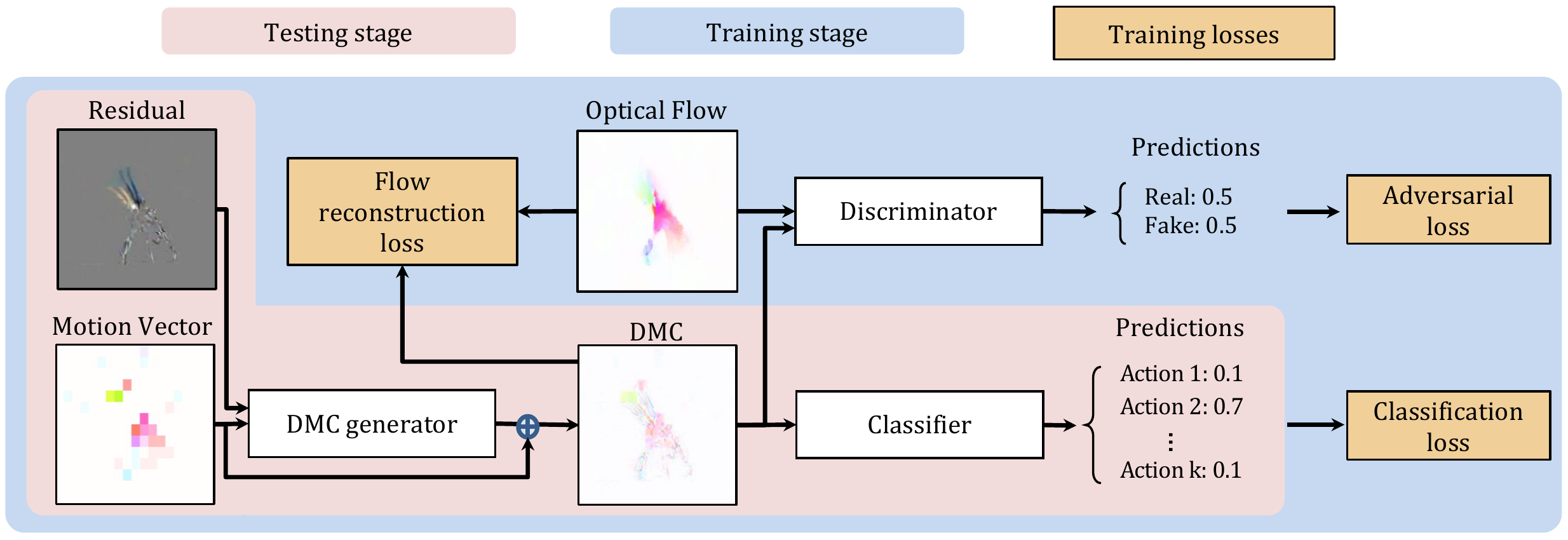}
\end{center}
\caption{The framework of our \ours (\oursnickname). Given the stacked residual and motion vector as input, the DMC generator reduces noise in the motion vector and captures more fine motion details, outputting a more discriminative motion cue representation which is used by a small classification network to classify actions. In the training stage, we train the DMC generator and the action classifier jointly using three losses. In the test stage, only the modules highlighted in pink are used. 
 }
\label{fig:framework}
\end{figure*}

In this section, we present our approach for generating \textit{Discriminative Motion Cues (\textbf{DMC})} from compressed video.
The overall framework of our proposed \textbf{\oursnickname} is illustrated in Figure~\ref{fig:framework}.
In Section~\ref{sec:compress video}, we introduce the basics of compressed video and the notations we use.
Then we design the DMC generator network in Section~\ref{sec:generator}.
Finally we present the training objectives in Section~\ref{sec:supervision} and discuss inference in Section~\ref{sec:inference}.

\subsection{Basics and Notations of Compressed Video}
\label{sec:compress video}
We follow
CoViAR \cite{wu2018coviar} and use MPEG-4 Part2 \cite{le1991mpeg} encoded videos where every I-frame is followed by 11 consecutive P-frames.
Three data modalities are readily available in MPEG-4 compressed video:
(1) RGB image of I-frame (\textbf{I});
(2) Motion Vector (\textbf{MV}) records the displacement of each macroblock in a P-frame to its reference frame and typically a frame is divided into 16x16 macroblocks during video compression;
(3) Residual (\textbf{R}) stores the RGB difference between a P-frame and its reference I-frame after motion compensation based on MV.
For a frame of height $H$ and width $W$, I and R have shape (3, $H$, $W$) and MV has shape (2, $H$, $W$).
But note that MV has much lower resolution in effect because its values within the same macroblock are identical.
\subsection{The Discriminative Motion Cue Generator}
\label{sec:generator}

\head{Input of the generator}.
Existing compressed video based methods directly feed motion vectors into a classifier to model motion information.
This strategy is not effective in modeling motion due to the characteristics of MV:
(1) MV is computed based on simple block matching, making MV noisy and (2) MV has substantially lower resolution, making MV lacking fine motion details.
In order to specifically handle these characteristics of MV, we aim to design a lightweight generation network to reduce noise in MV and capture more fine motion details, outputting DMC as a more discriminative motion representation.

To accomplish this goal, MV alone may not be sufficient.
According to \cite{sevilla2017integration}, the motion nearby object boundary is more important than the motion at other locations for action recognition.
We also notice R is often well-aligned with the boundary of moving objects.
Moreover, R is strongly correlated with MV as it is computed as the difference between the original frame and its reference I-frame compensated using MV.
Therefore, we propose to stack MV and R as input into the DMC generator, as shown in Figure \ref{fig:framework}. This allows utilizing the motion information in MV and R as well as the correlation between them, which cannot be modeled by separate CNNs as in the current compressed video works \cite{wu2018coviar,zhang2018real_DTMV,ZhangWWQW16_MVCNN}.

\begin{table}[t]
\begin{center}
\footnotesize
\begin{tabular}{c|c}
Network Architecture & GFLOPs \\ 
\specialrule{.2em}{.1em}{.1em}
C3D \cite{tran2015learning} & 38.5 \\
Res3D-18 \cite{tran2017convnet}  & 19.3 \\
ResNet-152 \cite{He_2016_CVPR} & 11.3 \\
ResNet-18 \cite{He_2016_CVPR} & 1.78 \\ \hline
DMC generator (PWC-Net \cite{Sun2018PWC-Net})  & 36.15 \\
DMC generator [ours] & 0.23
\end{tabular}
\end{center}
\vspace{-1em}
\caption{Computational complexity of different networks. Input has height 224 and width 224.} 
\label{table:complexity}
\end{table}

\begin{table}[t]\footnotesize
\begin{center}
\begin{tabular}{cccc}
Layer & Input size & Output size & Filter config \\ \hline
$\tt conv0$ & \textcolor{blue}{5}, 224, 224 & \textcolor{blue}{8}, 224, 224 & \textcolor{blue}{8}, 3x3, 1, 1 \\
$\tt conv1$ & \textcolor{blue}{13}, 224, 224 & \textcolor{blue}{8}, 224, 224 & \textcolor{blue}{8}, 3x3, 1, 1 \\
$\tt conv2$ & \textcolor{blue}{21}, 224, 224 & \textcolor{blue}{6}, 224, 224 & \textcolor{blue}{6}, 3x3, 1, 1 \\
$\tt conv3$ & \textcolor{blue}{27}, 224, 224 & \textcolor{blue}{4}, 224, 224 & \textcolor{blue}{4}, 3x3, 1, 1 \\
$\tt conv4$ & \textcolor{blue}{31}, 224, 224 & \textcolor{blue}{2}, 224, 224 & \textcolor{blue}{2}, 3x3, 1, 1 \\
$\tt conv5$ & \textcolor{blue}{33}, 224, 224 & \textcolor{blue}{2}, 224, 224 & \textcolor{blue}{2}, 3x3, 1, 1
\end{tabular}
\end{center}
\vspace{-1em}
\caption{The architecture of our Discriminative Motion Cue (DMC) generator network which takes stacked motion vector and residual as input. Input/output size follows the format of \#channels, height, width. Filter configuration follows the format of \#filters, kernel size, stride, padding.
}
\label{table:layers}
\end{table}

\hfill\break\head{Generator network architecture}. 
Quite a few deep generation networks have been proposed for optical flow estimation from RGB images. One of these works is PWC-Net~\cite{Sun2018PWC-Net}, which achieves SoTA performance in terms of both End Point Error (EPE) and inference speed.
We therefore choose to base our generator design principles on the ones used by PWC-Net.
It is worth noting that PWC-Net takes decoded RGB frames as input unlike our proposed method operating only in the compressed domain.

Directly adopting the network architecture of the flow estimator network in PWC-Net for our DMC generator leads to high GFLOPs as indicated in Table \ref{table:complexity}.
To achieve high efficiency, we have conducted detailed architecture search experimentally to reduce the number of filters in each convolutional layer of the flow estimator network in PWC-Net, achieving the balance between accuracy and complexity.
Furthermore, since our goal is to refine MV, we propose to add a shortcut connection between the input MV and the output DMC, making the generator to directly predict the refinements which are added on MV to obtain DMC.

Table \ref{table:layers} shows the network architecture of our DMC generator: 6 convolutional layers are stacked sequentially with all convolutional layers densely connected~\cite{huang2017densely}.
Every convolutional filter has a 3x3 kernel with stride 1 and padding 1.
Each convolutional layer except $\tt conv5$ is followed by a Leaky ReLU~\cite{maas2013rectifier} layer, where the negative slope is 0.1.

As shown in Table \ref{table:complexity}, our DMC generator only requires 0.63\% GFLOPs used by the flow estimator in PWC-Net if it were adopted to implement our DMC generator.
Also, Table \ref{table:complexity} compares our DMC generator with other popular network architectures for video analysis including frame-level models (ResNet-18 and ResNet-152 \cite{He_2016_CVPR}) and clip-level models (C3D \cite{tran2015learning} and Res3D~\cite{tran2017convnet}).
We observe that the complexity of DMC generator is orders of magnitude smaller compared to that of other architectures, which makes it running much faster.
In the supplementary material, we explored a strategy of using two consecutive networks to respectively rectify errors in MV and capture fine motion details while this did not achieve better accuracy.


\subsection{Flow-guided, Discriminative Motion Cues}
\label{sec:supervision}

Compared to MV, optical flow exhibits more discriminative motion information because: (1) Unlike MV is computed using simple block matching, nowadays dense flow estimation is computed progressively from coarse scales to fine scales \cite{TVL1}. (2) Unlike MV is blocky and thus misses fine details, flow keeps the full resolution of the corresponding frame.
Therefore we propose to guide the training of our DMC generator using optical flow.
To this end, we have explored different ways and identified three effective training losses as shown in Figure~\ref{fig:framework} to be presented in the following: a flow reconstruction loss, an adversarial loss, and a downstream classification loss.

\subsubsection{Optical Flow Reconstruction Loss}
First, we minimize the per-pixel difference between the generated DMC and its corresponding optical flow.
Following Im2Flow~\cite{gaoim2flow} which approximates flow from a single RGB image, we use the Mean Square Error (MSE) reconstruction loss ${{\cal L}_{\rm{mse}}}$ defined as:
\begin{equation}
\label{eq:flow recon}
{{\cal L}_{{\rm{mse}}}} = \mathbb{E}_{\vx \sim p} \left\| \mathcal{G}_{\rm{DMC}}(\vx) - \mathcal{G}_{\rm{OF}}(\vx) \right\|_2^2,
\end{equation}
where $p$ denotes the set of P-frames in the training videos, $\mathbb{E}$ stands for computing expectation, $\mathcal{G}_{\rm{DMC}}(\vx)$ and $\mathcal{G}_{\rm{OF}}(\vx)$\footnote{We relax the notational rigor and use $\mathcal{G}_{OF}(\vx)$ to refer to the optical flow corresponding to the frame $\vx$, although for many optical flow algorithms the input would be a pair of frames.} respectively denote the DMC and optical flow for the corresponding input frame $\vx$ sampled from $p$.
Since only some regions of flow contain discriminative motion cues that are important for action recognition, in the supplementary material we have explored weighting the flow reconstruction loss to encourage attending to the salient regions of flow. But this strategy does not achieve better accuracy.

\subsubsection{Adversarial Loss}\label{sec:gan}
As pointed out by previous works~\cite{mathieu2015deep}, the MSE loss implicitly assumes that the target data is drawn from a Gaussian distribution and therefore tends to generate smooth and blurry outputs.
This in effect results in less sharp motion representations especially around boundaries, making the generated DMC less discriminative.
Generative Adversarial Networks (GAN)~\cite{goodfellow2014generative} has been proposed to minimize the Jensen$-$Shannon divergence between the generative model and the true data distribution, making these two similar.
Thus in order to help our DMC generator learn to approximate the distribution of optical flow data, we further introduce an adversarial loss.
Note that unlike GAN which samples from random noise, adversarial loss samples from the input dataset, which
already has large variability~\cite{mathieu2015deep}.

Let our DMC generator $\mathcal{G}_{\rm{DMC}}$ be the \textbf{Generator} in the adversarial learning process. As shown in Figure \ref{fig:framework}, a \textbf{Discriminator $\mathcal{D}$} is introduced to compete with \textbf{$\mathcal{G}_{\rm{DMC}}$}. \textbf{$\mathcal{D}$} is instantiated by a binary classification network that takes as input either \textbf{real} optical flow or \textbf{fake} samples generated via our DMC generator.
Then $\mathcal{D}$ outputs a two-dimensional vector that is passed through a softmax operation to obtain the probability $P_\mathcal{D}$ of the input being \textit{Real}, \ie flow versus \textit{Fake}, \ie DMC.
$\mathcal{G}_{\rm{DMC}}$ and $\mathcal{D}$ are trained in an alternating manner: $\mathcal{G}_{\rm{DMC}}$ is fixed when $\mathcal{D}$ is being optimized, and vice versa.

During training $\mathcal{D}$, $\mathcal{G}_{\rm{DMC}}$ is fixed and is only used for inference. $\mathcal{D}$ aims to classify the generated DMC as Fake and classify flow as Real.
Thus the adversarial loss for training $\mathcal{D}$ is:
\begin{equation}
\label{eq:d}
\begin{split}
{{\cal L}_{\rm{adv}}^D} =  & \mathbb{E}_{\vx \sim p} [ - \log P_{\mathcal{D}} ( \rm{Fake} | \mathcal{G}_{\rm{DMC}}(\vx)) \\
& - \log P_{\mathcal{D}} (\rm{Real} | \mathcal{G}_{\rm{OF}}(\vx)) ],
\end{split}
\end{equation}
where $p$ denotes the set of P-frames in the training set and  $\mathcal{G}_{\rm{DMC}}(\vx)$ and $\mathcal{G}_{\rm{OF}}(\vx)$ respectively represent the DMC and optical flow for each input P-frame $\vx$.

During training $\mathcal{G}_{\rm{DMC}}$, $\mathcal{D}$ is fixed.
$\mathcal{G}_{\rm{DMC}}$ is encouraged to generate DMC that is similar and indistinguishable with flow.
Thus the adversarial loss for training $\mathcal{G}_{\rm{DMC}}$ is:
\begin{equation}
\label{eq:g}
{{\cal L}_{\rm{adv}}^G} =  \mathbb{E}_{\vx \sim p} [ - \log P_{\mathcal{D}} ( \rm{Real} | \mathcal{G}_{\rm{DMC}}(\vx))],
\end{equation}
which can be trained jointly with the other losses designed for training the DMC generator in an end-to-end fashion, as presented in Section~\ref{sec:loss}.

Through the adversarial training process, \textbf{$\mathcal{G}_{\rm{DMC}}$} learns to approximate the distribution of flow data, generating DMC with more fine details and thus being more similar to flow.
Those fine details usually capture discriminative motion cues and are thus important for action recognition.
We present details of the discriminator network architecture in the supplementary material.

\subsubsection{The Full Training Objective Function}
\label{sec:loss}

\head{Semantic classification loss}.
As our final goal is to create motion representation that is discriminative with respect to the downstream action recognition task, it is important to train the generator jointly with the follow-up action classifier.
We employ the softmax loss as our action classification loss, denoted as ${{\cal L}_{\rm{cls}}}$. 

\begin{figure*}[t]
\vspace{-2em}
\center
\includegraphics[width=\linewidth]{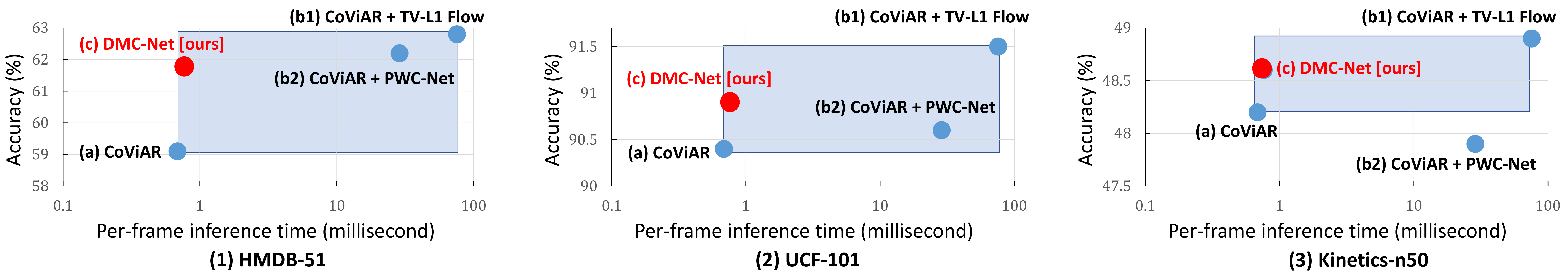}
\caption{Accuracy vs. speed on 3 benchmarks. Results on UCF-101 and HMDB-51 are averaged over 3 splits.
(b1) and (b2) use ResNet-18 to classify flow and (c) also uses ResNet-18 to classify DMC.
The proposed \textit{DMC-Net} not only operates exclusively in the compressed domain, but also is able to achieve higher accuracy than (a) while being two orders of magnitude faster than methods that use optical flow.
The blue area indicates the improvement room from (a) to (b1).}
\label{fig:res_sota}
\end{figure*}

\head{The full training objective}.
Our whole model is trained with the aforementioned losses putting together in an end-to-end manner.
The training process follows the alternating training procedure stated in Section~\ref{sec:gan}.
During training the discriminator, $\mathcal{D}$ is trained while the DMC generator $\mathcal{G}_{\rm{DMC}}$ and the downstream action classifier are fixed.
The full training objective is to minimize the adversarial loss ${{\cal L}_{\rm{adv}}^D}$ in Equation \ref{eq:d}.
During training the generator $\mathcal{G}_{\rm{DMC}}$, $\mathcal{D}$ is fixed while the DMC generator $\mathcal{G}_{\rm{DMC}}$ and the downstream action classifier are trained jointly with the following full training objective to be minimized:
\begin{equation}
\label{eq:total gen}
{{\cal L}_{{\rm{cls}}}} + \alpha  \cdot {{\cal L}_{{\rm{mse}}}}
 + \lambda  \cdot {{\cal L}_{\rm{adv}}^G},
\end{equation}
where ${{\cal L}_{{\rm{mse}}}}$ is given by Equation \ref{eq:flow recon}, ${{\cal L}_{\rm{adv}}^G}$ is given by Equation \ref{eq:g}, and  $\alpha, \lambda$ are balancing weights. 




\subsection{Inference} 
\label{sec:inference}

As shown in Figure \ref{fig:framework}, despite having three losses jointly trained end-to-end, our \oursnickname is actually quite efficient during inference: basically first the generator outputs DMC and then the generated DMC is fed into the classification network to make action class prediction. 
We compare our inference speed with other methods in Section~\ref{sec:speed}.



\section{Experiments}
\label{sec:experiments}

In this section, we first detail our experimental setup, present quantitative analysis of our model, and finally compare with state-of-the-art methods.


\subsection{Datasets and Evaluation}

\noindent\textbf{UCF-101 \cite{UCF101}}. This dataset contains 13,320 videos from 101 action categories, along with 3 public train/test splits.

\noindent\textbf{HMDB-51 \cite{kuehne2011hmdb}}.
This dataset contains 6,766 videos from 51 action categories, along with 3 public train/test splits.

\noindent\textbf{Kinetics-n50}.
From the original Kinetics-400 dataset \cite{Kinetics}, we construct a subset referred as \textbf{Kinetics-n50} in this paper. We keep all 400 categories. For each class, we randomly sample 30 videos from the original training set as our training videos and randomly sample 20 videos from the original validation set as our testing videos. We evaluate on the full set in the supplementary material.

\noindent\textbf{Evaluation protocol}.
All videos in the above datasets have single action label out of multiple classes. Thus we evaluate top-1 video-level class prediction accuracy. 

\subsection{Implementation Details}\label{sec:implementation}

\noindent\textbf{Training}.
For I, MV, and R, we follow the exactly same setting as used in CoViAR \cite{wu2018coviar}.
Note that I employs ResNet-152 classifier; MV and R use ResNet-18 classifier.
To ensure efficiency, \oursnickname also uses ResNet-18 to classify DMC in the whole paper unless we explicitly point out.
To allow apple-to-apple comparisons between DMC and flow, we also choose frame-level ResNet-18 classifier as the flow CNN shown in Figure \ref{fig:sub2}.
TV-L1 \cite{zach2007duality} is used for extracting optical flow to guide the training of our \oursnickname.
All videos are resized to 340$\times$256. Random cropping of 224$\times$224 and random flipping are used for data augmentation.
More details are in the supplementary material.

\noindent\textbf{Testing}.
For I, MV, and R, we follow the exactly same setting as in CoViAR \cite{wu2018coviar}: 25 frames are uniformly sampled for each video; each sampled frame has 5 crops augmented with flipping; all 250 (25$\times$2$\times$5) score predictions are averaged to obtain one video-level prediction. 
For DMC, we following the same setting except that we do not use cropping and flipping, which shows comparable accuracy but requires less computations.
Finally, we follow CoViAR \cite{wu2018coviar} to obtain the final prediction via fusing prediction scores from all modalities (\ie I, MV, R, and DMC).

\begin{table*}[t]
\footnotesize
\center
\begin{tabular}{cc|cc|cc}
 &  & \multicolumn{2}{c|}{\begin{tabular}[c]{@{}c@{}}Two-Stream Method \\ (RGB+Flow) \end{tabular}} & \multicolumn{2}{c}{\begin{tabular}[c]{@{}c@{}}Compressed Video \\ Based Methods \end{tabular}} \\ \cline{3-6} 
 &  & BN-Inception  & ResNet152 & CoViAR  & \oursnickname [ours] \\ \hline
\multicolumn{1}{c|}{\multirow{5}{*}{\begin{tabular}[c]{@{}c@{}}Time \\ (ms) \end{tabular}}} & Preprocess & 75.0 & 75.0 & 0.46 & 0.46 \\ \cline{2-6} 
\multicolumn{1}{c|}{} & CNN (\textbf{S}) & 1.6 & 7.5 & 0.59 & 0.89 \\
\multicolumn{1}{c|}{} & Total (\textbf{S}) & 76.6 & 82.5 & 1.05 & 1.35 \\ \cline{2-6} 
\multicolumn{1}{c|}{} & CNN (\textbf{C}) & 0.9 & 4.0 & 0.22 & 0.30 \\
\multicolumn{1}{c|}{} & Total (\textbf{C}) & 75.9 & 79.0 & 0.68 & 0.76 \\ \hline
\multicolumn{1}{c|}{\multirow{2}{*}{FPS}} & CNN (\textbf{C}) & 1111.1 & 250.0 & 4545.4 & 3333.3 \\
\multicolumn{1}{c|}{} & Total (\textbf{C}) & 13.1 & 12.6 & 1470.5 & 1315.7 \\
\multicolumn{6}{c} {} \\
\multicolumn{6}{c}{(a) \oursnickname vs. Two-stream methods and CoViAR}
\end{tabular}
\begin{tabular}{c|c|c}
 & Generator & Generator + Cls. \\ \cline{2-3} 
 & Time (ms) / FPS & Time (ms) / FPS \\ \hline
Deepflow \cite{weinzaepfel2013deepflow} &  1449.2 / 0.7 & 1449.5 / 0.7 \\
Flownet2.0 \cite{ilg2017flownet} & 220.8 / 4.5 & 221.0 / 4.5 \\
TVNet \cite{fan2018end} & 83.3 / 12.0 & 83.5 / 12.0 \\
PWC-Net \cite{Sun2018PWC-Net} & 28.6 / 35.0 & 28.8 / 34.8 \\
\textbf{\oursnickname [ours]} & \textbf{0.1} / \textbf{9433.9} & \textbf{0.3} / \textbf{3333.3} \\
\multicolumn{3}{c}{} \\
\multicolumn{3}{c}{(b) \oursnickname vs. flow estimation methods}
\end{tabular}
\caption{Comparisons of per-frame inference speed.
\textbf{(a)} Comparing our \oursnickname to the two-stream methods~\cite{ioffe2015batch, He_2016_CVPR} and the CoViAR method~\cite{wu2018coviar}.  We consider two scenarios of forwarding multiple CNNs sequentially and concurrently, denoted by \textbf{S} and \textbf{C} respectively.
We measure CoViAR's CNN forwarding time using our own implementation as mentioned in Section \ref{sec:speed} and numbers are comparable to those reported in \cite{wu2018coviar}.
\textbf{(b)} Comparing our \oursnickname to deep network based optical flow estimation and motion representation learning methods, whose numbers are quoted from \cite{fan2018end}. CNNs in \oursnickname are forwarded concurrently. All networks have batch size set to 1.
For the classifier (denoted as Cls.), all methods use ResNet-18.
}
\label{table:speed}
\end{table*}

\begin{figure*}[t]
\center
\includegraphics[width=.8\linewidth]{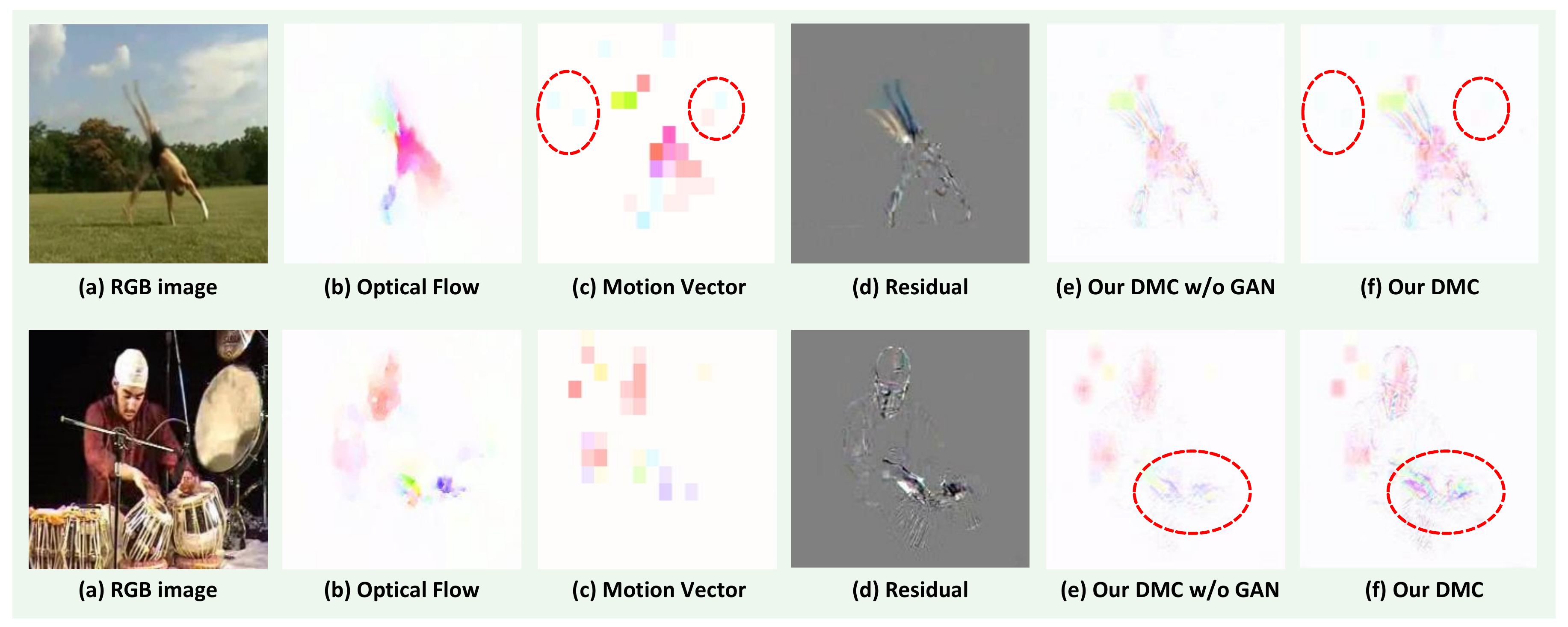}
\caption{A Cartwheel example (top) and a PlayingTabla (bottom) example. All images in one row correspond to the same frame.
For the Cartwheel example, these noisy blocks in the background (highlighted by two red circles) are reduced in our DMC.
For the PlayingTabla example, our DMC exhibits sharper and more discriminative motion cues around hands (highlighted by the red circle) than our DMC w/o the adversarial loss during training.
Better viewed in color.}
\label{fig:vis}
\end{figure*}

\begin{table}[t]
\footnotesize
\center
\begin{tabular}{lcc}
\multicolumn{1}{c}{} & \multicolumn{1}{c}{HMDB-51} & \multicolumn{1}{c}{UCF-101} \\ \hline
\multicolumn{3}{l}{\textbf{Compressed video based methods}} \\
 &  &  \\
EMV-CNN \cite{ZhangWWQW16_MVCNN} & 51.2 (split1) & 86.4 \\
DTMV-CNN \cite{zhang2018real_DTMV} & 55.3 & 87.5 \\
CoViAR \cite{wu2018coviar} & 59.1 & 90.4 \\
\textbf{\oursnickname (ResNet-18) [ours]} & 62.8 & 90.9 \\ 
\textbf{\oursnickname (I3D) [ours]} & 71.8 & 92.3 \\ 
&  & \\ \hline
\multicolumn{3}{l}{\textbf{Decoded video based methods} \textbf{\textit{(RGB only)}}} \\
\multicolumn{2}{c}{\textit{\textbf{Frame-level classification}}} & \\
ResNet-50 \cite{He_2016_CVPR} & 48.9 & 82.3 \\
ResNet-152 \cite{He_2016_CVPR} & 46.7 & 83.4 \\
\multicolumn{2}{c}{\textit{\textbf{Motion representation learning}}} & \\
ActionFlowNet (2-frames) \cite{ng2018actionflownet} & 42.6 & 71.0 \\
ActionFlowNet \cite{ng2018actionflownet} & 56.4 & 83.9 \\ 
PWC-Net (ResNet-18) + CoViAR \cite{Sun2018PWC-Net} & 62.2 & 90.6 \\
TVNet \cite{fan2018end} & 71.0 & 94.5 \\
\multicolumn{2}{c}{\textit{\textbf{Spatio-temporal modeling}}} & \\
C3D \cite{tran2015learning} & 51.6 & 82.3 \\
Res3D \cite{tran2017convnet} & 54.9 & 85.8 \\
ARTNet \cite{wang2017appearance} & 70.9 & 94.3 \\ 
MF-Net \cite{chen2018multi} & 74.6 & 96.0 \\
S3D \cite{xie2017rethinking} & 75.9 & 96.8 \\ 
I3D RGB \cite{Kinetics} & 74.8 & 95.6 \\
\textbf{I3D RGB + \oursnickname (I3D) [ours]} & 77.8 & 96.5 \\
 &  & \\ \hline
\multicolumn{3}{l}{\textbf{Decoded video based methods} \textbf{\textit{(RGB + Flow)}}} \\
Two-stream \cite{Simonyan14b} & 59.4 & 88.0 \\
Two-Stream fusion \cite{feichtenhofer2016convolutional} & 65.4 & 92.5 \\
I3D \cite{Kinetics} & 80.7 & 98.0 \\
R(2+1)D \cite{tran2018closer} & 78.7 & 97.3
\end{tabular}
\caption{Accuracy averaged over all three splits on HMDB-51 and UCF-101 for both state-of-the-art compressed video based methods and decoded video based methods.}
\label{table:res_decoded}
\end{table}

\subsection{Model Analysis}\label{sec:discussions}\label{sec:explore}


\head{How much gain \textbf{\oursnickname} can improve over \textbf{CoViAR}?}
Figure \ref{fig:res_sota} reports accuracy on all three datasets.
\textbf{CoViAR + TV-L1} and \textbf{CoViAR + PWC-Net} follow two-stream methods to include an optical flow stream computed by TV-L1 \cite{TVL1} and PWC-Net \cite{Sun2018PWC-Net} respectively.
\textbf{CoViAR + TV-L1} can be regard as our upper bound for improving accuracy because TV-L1 flow is used to guide the training of \textbf{\oursnickname}.
By only introducing a lightweight DMC generator, our \textbf{\oursnickname} significantly improves the accuracy of \textbf{CoViAR} to approach \textbf{CoViAR + Flow}.
Figure \ref{fig:vis} shows that the generated DMC has less noisy signals such as those in the background area and DMC captures fine and sharp details of motion boundary, leading to the accuracy gain over \textbf{CoViAR}.

\head{How effectiveness is each proposed loss?}
On HMDB-51, when only using the classification loss, the accuracy of \oursnickname is 60.5\%; when using the classification loss and the flow reconstruction loss, the accuracy is improved to 61.5\%; when further including the adversarial training loss, \oursnickname eventually achieves 61.8\% accuracy. 
As indicated by previous literature~\cite{isola2017image}, using an adversarial loss without a reconstruction loss often introduces artifacts.

\subsection{Inference Speed}\label{sec:speed}

Following \cite{wu2018coviar}, we measure the average per-frame running time, which consists of the time for data pre-processing and the time for CNN forward pass.
For the CNN forward pass, both the scenarios of forwarding multiple CNNs sequentially and concurrently are considered.
Detailed results can be found in Table \ref{table:speed} (a).
Results of \textbf{two-stream methods} are quoted from \cite{wu2018coviar}.
Due to the need of decoding compressed video into RGB frames and then computing optical flow, its pre-process takes much longer time than compressed video based methods.
\textbf{\oursnickname} accepts the same inputs as \textbf{CoViAR} and thus \textbf{CoViAR} and \textbf{\oursnickname} have the same pre-processing time.
As for the CNN forwarding time of compressed video based methods, we measure \textbf{CoViAR} and \textbf{\oursnickname} using the exactly same implementation as stated in Section \ref{sec:implementation} and the same experimental setup: we use one NVIDIA GeForce GTX 1080 Ti and set the batch size of each CNN to 1 while in practice the speed can be further improved to utilize larger batch size.
Despite adding little computational overhead on CoViAR, \oursnickname is still significantly faster than the conventional \textbf{two-stream methods}.

\textbf{Deepflow} \cite{weinzaepfel2013deepflow}, \textbf{Flownet} \cite{ilg2017flownet} and \textbf{PWC-Net} \cite{Sun2018PWC-Net} have been proposed to accelerate optical flow estimation by using deep networks.
\textbf{TVNet} \cite{fan2018end} was proposed to generate even better motion representation than flow with fast speed.
Those estimated flow or generated motion representation can replace optical flow used in two-stream methods to go through a CNN for classification.
We combine these methods with a ResNet-18 classifier in Table \ref{table:speed} (b). We can see that our DMC generator runs much faster than these state-of-the-art motion representation learning methods.

\subsection{Comparisons with Compressed Video Methods}


As shown in the top section of Table \ref{table:res_decoded}, \textbf{\oursnickname} outperforms all other methods that operate in the compressed video domain, \ie \textbf{CoViAR}~\cite{wu2018coviar}, \textbf{EMV-CNN}~\cite{ZhangWWQW16_MVCNN} and \textbf{DTMV-CNN}~\cite{zhang2018real_DTMV}. Our method outperforms methods like~\cite{ZhangWWQW16_MVCNN, zhang2018real_DTMV} that the output of the MV classifier is trained to approximate the output of the optical flow classifier. We believe this is because of the fact that approximating the classification output directly is not ideal, as it does not explicitly address the issues that MV is noisy and low-resolutional. By generating a more discriminative motion representation DMC, we are able to get features that are highly discriminative for the downstream recognition task.
Furthermore, our \oursnickname can be combined with these classification networks of high capacity and trained in an end-to-end manner.
\textbf{DMC-Net (I3D)} replaces the classifier from ResNet-18 to I3D, achieving significantly higher accuracy and outperforming a number of methods that require video decoding.
Our supplementary material discusses the speed of I3D.

\subsection{Comparisons with Decoded Video Methods}\label{sec:decoded}
In this section we compare \oursnickname to approaches that require decoding all RGB images from compressed video. Some only use the RGB images, while others adopt the two-stream method~\cite{Simonyan14b} and further require computing flow.

\noindent\textbf{RGB only}.
As shown in Table \ref{table:res_decoded}, decoded video methods only based on RGB images can be further divided into three categories.
\textbf{(1) Frame-level classification}: 2D CNNs like ResNet-50 and ResNet-152 \cite{He_2016_CVPR} have been experimented in \cite{feichtenhofer2017spatiotemporal} to classify each frame individually and then employ simple averaging to obtain the video-level prediction.
Due to lacking motion information, frame-level classification underperforms \textbf{\oursnickname}.
\textbf{(2) Motion representation learning}:
In Table \ref{table:res_decoded}, we evaluate \textbf{PWC-Net (ResNet-18) + CoViAR} which feeds estimated optical flow into a ResNet-18 classifier and then fuses the prediction with \textbf{CoViAR}.
The accuracy of \textbf{PWC-Net (ResNet-18) + CoViAR} is not as good as \oursnickname because our generated DMC contains more discriminative motion cues that are complementary to MV.
For TVNet \cite{fan2018end}, the authors used BN-Inception \cite{ioffe2015batch} to classify the generated motion representation and then fuse the prediction with a RGB CNN.
The accuracy of TVNet is better \textbf{DMC-Net (ResNet-18)} thanks to using a strong classifier but is worse than our \textbf{DMC-Net (I3D)}.
\textbf{(3) Spatio-temporal modeling}: There are also a lot of works using CNN to model the spatio-temporal patterns across multiple RGB frames to implicitly capture motion patterns.
It turns out that our \textbf{\oursnickname} discovers motion cues that are complementary to such spatio-temporal patterns: \textbf{I3D RGB + \oursnickname (I3D)} improves \textbf{I3D RGB} via incorporating predictions from our \textbf{\oursnickname (I3D)}.

\noindent\textbf{RGB + Flow}.
As shown in Table \ref{table:res_decoded}, the state-of-the-art accuracy is belonging to the two-stream methods \cite{kay2017kinetics,tran2018closer}, which combine predictions made from a RGB CNN and an optical flow CNN.
But as discussed in Section \ref{sec:speed}, extracting optical flow is quite time-consuming and thus these two-stream methods are much slower than our \textbf{\oursnickname}.


\section{Conclusion}
\label{sec:conclusions}
In this paper, we introduce \textbf{DMC-Net}, a highly efficient deep model for video action recognition in the compressed video domain.
Evaluations on 3 action recognition benchmarks lead to substantial gains in accuracy over prior work, without the assistance of computationally expensive flow.
The supplementary materials can be found in the following appendix.



\section{Acknowledgment}

Zheng Shou would like to thank the support from Wei Family Private Foundation when Zheng was at Columbia.


\section{Appendix}

\subsection{Data Modalities in the Compressed Domain}

Prevailing video compression standards employs the Group Of Pictures (GOP) structure to encode the raw video into successive, non-overlapping GOPs. Frames or pictures within one GOP are compressed together.
Each GOP begins with an I-frame (intra coded frame) whose RGB pixel values are stored.
I-frame can be decoded independently with other frames.

The rest of frames within a GOP are P-frame (predictive coded frame) and/or B-frame (bi-predictive coded frame), containing motion-compensated difference information relative to the previously decoded frames.
Each P-frame can only reference one frame which could be either I-frame or P-frame while each B frame can only reference two frames.
In this thesis, we follow \cite{wu2018coviar} to focus on the low-latency scenario which only involves P-frame without B-frame.
Each P-frame stores motion vectors and residual errors:
during encoding, the video codec divides a P-frame into macroblocks of size such as 16x16 and find the most similar image patch in the reference frame for each macroblock;
the displacement between a macroblock in P-frame and its most similar image patch in the reference frame is regarded as the corresponding motion vector, which will be used in motion compensation during decoding;
the pixel differences between a macroblock in P-frame and its most similar image patch in the reference frame are denoted as residual errors.
During the decoding of a P-frame, the video codec performs motion compensation which effectively warps the reference frame using the motion vectors and then adds the residual errors to the motion-compensated reference frame to reconstruct the P-frame.

Consequently, three data modalities in the compression domain are available: (1) RGB values of I-frame; (2) motion vectors and (3) residual errors of P-frame. We refer readers to \cite{le1991mpeg} for more details.

\subsection{More implementation details}
In addition to Section 4.2 in the main paper, here we present more implementation details. Our model is implemented using PyTorch \cite{paszke2017automatic}. We first train our \oursnickname with the adversarial loss and then train it with all losses together. We elaborate these two steps separately in the following. On all three datasets (\ie HMDB-51, UCF-101 and Kinetics-n50), we found a generic settings can work well.
We use Adam optimizer \cite{kingma2014adam}.

\hfill\break
\head{Configuration of the training with the flow reconstruction loss and the classification loss}.
We first train the DMC generator for 1 epoch using the flow reconstruction loss only with the classification network fixed.
Then we include the classification loss to train both the generator and classifier end-to-end for 49 epochs.
In the total loss (\ie the Equation 4 in the main paper), we set $\alpha$ to 10 to balance weights.
The overall learning rate is set to 0.01 and it is divided by 10 whenever the total training loss plateaus.
All layers in the classification network except its last layer have the learning rate set to be 100x smaller.

\hfill\break
\head{Configuration of the training with all losses including the adversarial loss}.
Then we use the above trained model as the initialization for training our whole model with all three losses including the adversarial loss.
Our whole model consists of the generator, the classifier and the discriminator now.
In the total loss (\ie the Equation 4 in the main paper), we set $\alpha$ to 10 and set $\lambda$ to 1.
The overall learning rate is set to 0.01 and it is divided by 10 whenever the total training loss plateaus.
All layers in the classification network except its last layer have the learning rate set to be 100x smaller.
Based on the network architectures for the discriminator used in a popular GAN implementation repository \footnote{https://github.com/eriklindernoren/PyTorch-GAN/tree/master/implementations}, we experimented with various number of filters in each layer and various number of layers.
Finally we identified a network architecture for implementing our discriminator which achieves accuracy comparable to more complicated architectures.
This discriminator's architecture consists of a stack of 2D convolutional layers with a two-way Fully Connected layer at the end, as shown in the following Figure \ref{fig:discriminator}. 

\begin{figure}[h]
\centering
\centerline{\includegraphics[width=0.36\linewidth]{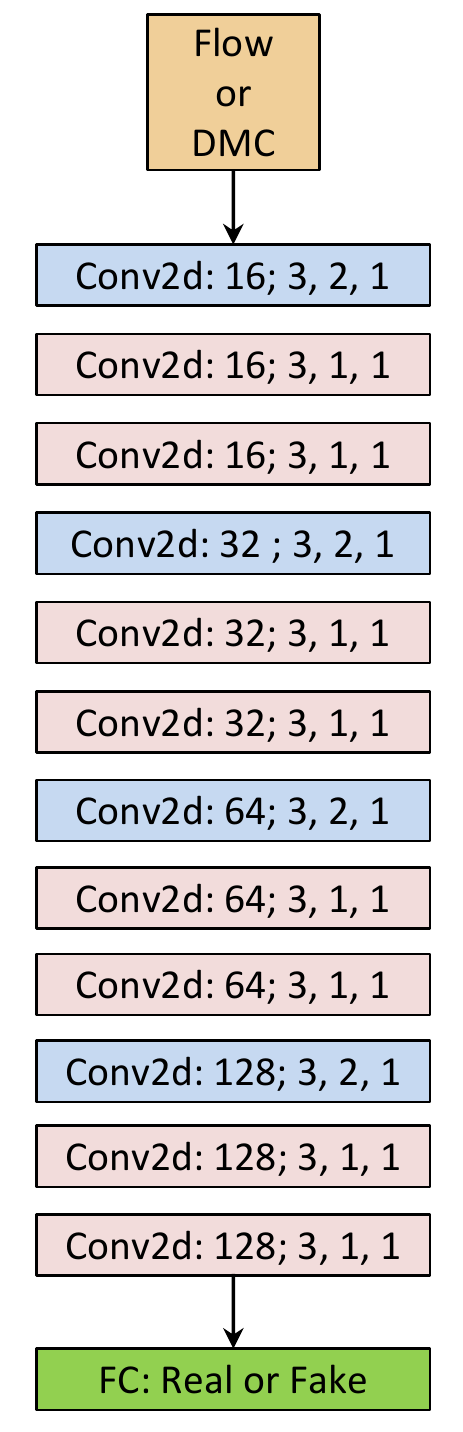}}
\caption{The network architecture of our discriminator. We denote each 2D convolutional layer in the format of \#filters; kernel size, stride, padding.}
\label{fig:discriminator}
\end{figure}

\subsection{Other early fusion possibilities}

\begin{figure}[h]
	\centering
	\centerline{\includegraphics[width=\linewidth]{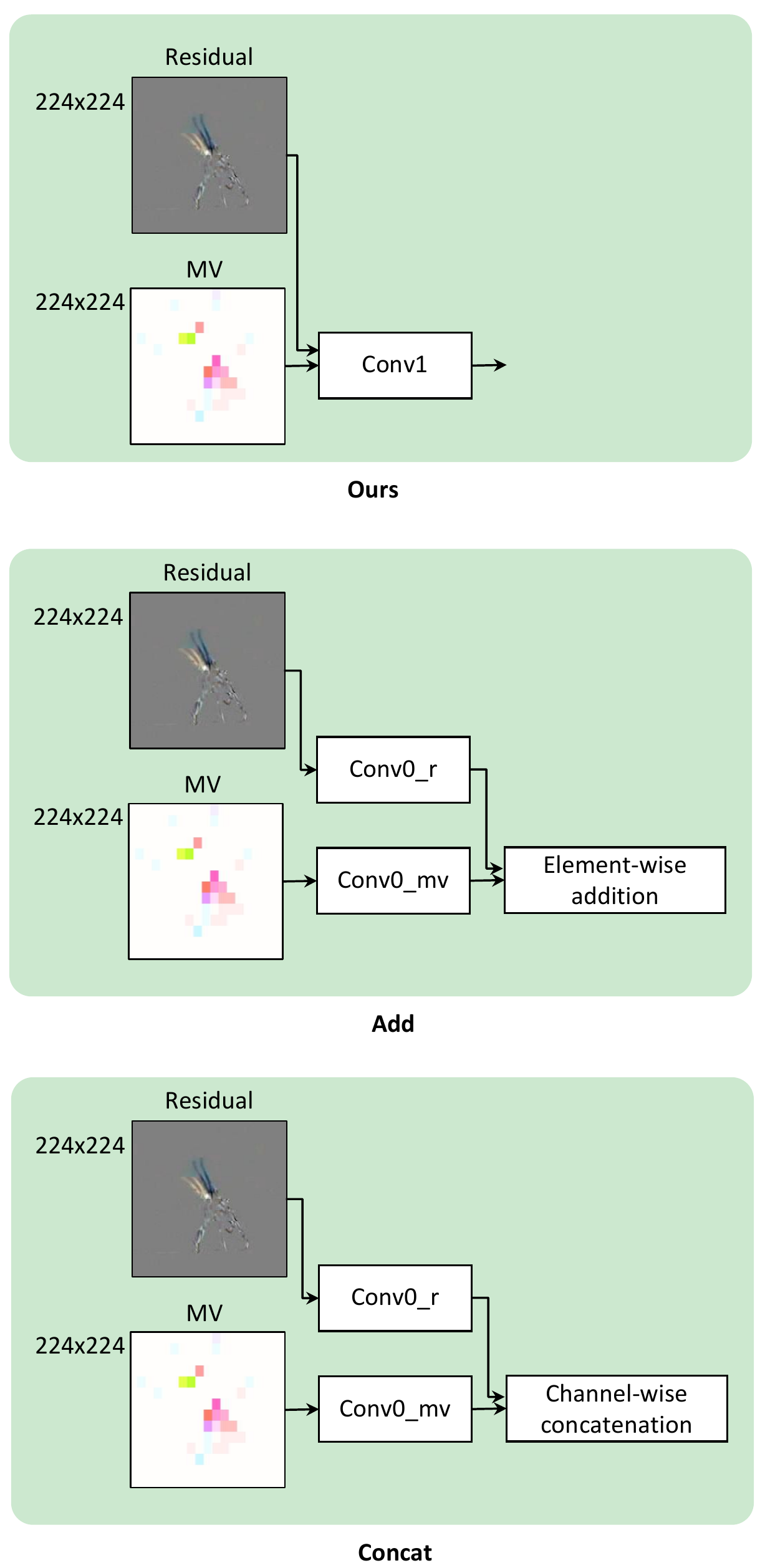}}
	\caption{Different early fusion possibilities.}
	\label{fig:earlyfusion}
\end{figure}

As shown in Figure~\ref{fig:earlyfusion}, we explore other early fusion possibilities:
we duplicate the first convolution layer (\textit{i.e.} $\tt conv0$) of our DMC generator as $\tt conv0$\_mv and $\tt conv0$\_r to respectively process MV and R independently.
Their outputs are fused before feeding into $\tt conv1$ and two fusion methods are studied: element-wise addition (denoted as \textbf{Add}) and channel-wise concatenation (denoted as \textbf{Concat}). 
On HMDB-51, our method (\textit{i.e.} directly stacking MV and R) achieves accuracy \textbf{61.80\%}, which is better than \textbf{Add} (61.32\%) and \textbf{Concat} (61.36\%).
We believe this is because MV and R are strongly correlated in the original pixel space before convolution.

\subsection{Attention-weighted flow reconstruction loss}

In this section we describe a way to attend to the discriminative regions of  optical flow during generating DMC.
However, in our experiments we found that this idea does not offer quantitative benefit beyond the GAN method on the datasets we experimented with.
Thus this idea was not included in the main paper.

\subsubsection{Approach}

The Mean Square Error (MSE) loss penalizes errors evenly over the whole image. In many cases, parts of optical flow contain noises, \eg motions corresponding to background or camera motion. When reconstructing the optical flow, our DMC generator would ideally focus only on parts of the flow that contain motion cues discriminative with respect to the downstream action recognition task.
Because these would be the regions of optical flow that are important for action recognition, and the regions where we would want a better reconstruction. Conversely, the reconstruction error in other regions of the optical flow, such as background, may not be important or even could be misleading.

This motivates us to try to create an adaptive MSE loss, where a weight is assigned for each location of the optical flow, based on the discriminative ability of that location. To get such a set of weights for each optical flow, we utilize recent related works on network interpretation~\cite{deconv_eccv14}, including the Class Activation Map~\cite{zhou2016learning} method and the Guided Back-Propagation~\cite{springenberg2014striving} method.
Such methods were proposed with a view to highlighting discriminative regions of the input data with respect to the classification outputs and are able to calculate \textit{attention}-like weights for every location of the input data.

All methods mentioned above require a trained classifier to inspect. We therefore first train a ResNet-18 classifier network for action recognition using optical flow as input.
We then use network interpretation methods to output a set of attention-like weights $A \in \mathbb{R}^{H \times W}$ for each input optical flow of height $H$ and width $W$. These attention-like weights can be computed before training our DMC generator and then be utilized during the training of our \oursnickname. Specifically, we can extend the optical flow reconstruction loss of Equation 1 in the main paper to take into account the location-specific weights and derive the adaptively weighted flow reconstruction loss ${{\cal L}_{{\rm{MSE-\alpha}}}}$:
\begin{equation}
\begin{split}
 {{\cal L}_{{\rm{MSE-\alpha}}}} &  = \mathbb{E}_{\vx \sim p} \frac{1}{H \times W} \\
&  \sum_{h = 1}^H \sum_{w = 1}^W
A_{h,w} \cdot \left\|  \phi_{h,w} - \sigma_{h,w} \right\|_2^2,
\end{split}
\end{equation}
where $p$ denotes the set of P-frames in the training videos, $\phi$ is set to the generated DMC denoted as $ {\mathcal{G}_{\rm{DMC}}}(\vx)$, $\sigma$ is set to the corresponding optical flow denoted as ${\mathcal{G}_{\rm{OF}}}(\vx)$, $\mathbb{E}$ stands for computing expectation, $A_{h,w}$ denotes the learned weight for location $h, w$. In order to obtain $A$, we have explored two widely used network interpretation techniques as presented in the following.

\begin{figure*}[t]
\center
\includegraphics[width=\linewidth]{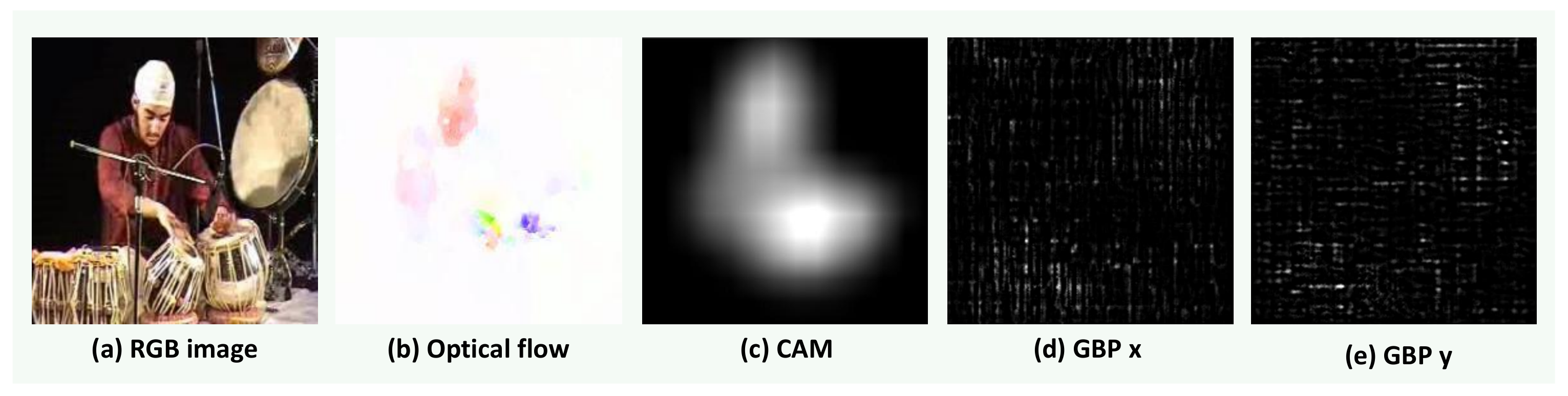}
\caption{
Illustrations of the attention maps generated by CAM \cite{zhou2016learning} and GBP \cite{springenberg2014striving} for a PlayingTabla example. (a) shows the RGB image. (b) shows the corresponding optical flow. (c) is the attention map generated using the CAM method. (d) and (e) show the attention maps generated using the GBP method for the input flow respectively along the x direction and the y direction.
Better viewed in color.}
\label{fig:vis_heatmap}
\end{figure*}

\hfill\break
\head{Class Activation Map (CAM)} \cite{zhou2016learning}.
The CAM method practically switches the order of the last two layers of the trained flow classifier, \ie the fully connected classification layer and the Global Average Pooling layer $\tt pool5$. This way, the fully connected classifier can be re-purposed as a convolutional layer $\tt f_{cls}$ to slide over every location of the $\tt conv5$ (\ie the layer right before the $\tt pool5$)'s output, effectively producing a classification score at each location.
As the output of $\tt f_{cls}$ is of low spatial dimension and each location has a wide receptive field with respect to the input flow, the high activations effectively focus on the most discriminative salient regions of the input flow.
We choose the activation map corresponding to the ground truth action class as the attention map $A$.
Finally we deal with the negative values in $A$ via passing $A$ through a ReLU operation, which leads to the best accuracy compared to other common normalization methods according to our experimental explorations.
Note that in our experiments discussed in the following Section \ref{sec:exp_att}, we resize the input flow from 224x224 to 448x448 before feeding it into the classifier so that we can obtain the attention map $A$ of higher spatial resolution (\ie 14x14), covering more details.
Further, we upsample $A$ back to 224x224 via bilinear interpolation so that $A$ has the same size as the generated DMC.
As shown in Figure~\ref{fig:vis_heatmap} (c), the attention map generated by the CAM method can indeed highlight the salient regions of flow such as the player's hands and head.
The flow values along the x direction and the y direction at the same spatial location share the same attention weight.

\hfill\break
\head{Guided Back-Propagation (GBP)} \cite{springenberg2014striving}.
Rather than finding the salient regions, some methods \cite{deconv_eccv14,springenberg2014striving} have been proposed to determine the contribution from the input's each value to the final classification output.
Since the input in our case is optical flow, the higher the contribution of a value, the more discriminative motion information this value contains.
Therefore, we can obtain an attention map $A$ of the data shape as the same as the flow (\ie 2x224x224 in the following Section \ref{sec:exp_att}).
Each value in $A$ stands for the contribution of the corresponding flow's value at the same location.
Specifically, we utilize the GBP~\cite{springenberg2014striving} method, which improves the De-conv method~\cite{deconv_eccv14} by combining it with the regular back-propagation pass.
Concretely, we set the classification output as a one-hot vector with the ground truth class indicated and then we back-propagate the one-hot vector back to the input flow.
Note that following the conventional back-propagation can only generate a generic attention map independent to the input rather than a map that is related with a specific input flow.
To address this issue, GBP further integrates the De-conv method into the conventional back-propagation pass: basically whenever back-propagating gradients through a ReLU layer, GBP sets the negative gradients to 0.
Finally, we pass the obtained $A$ through a ReLU operation to set its negative values to 0.
Figure~\ref{fig:vis_heatmap} (d) and (e) show the attention maps generated by the GBP method, highlighting the pixels whose values are sensitive for classifying the input optical flow as PlayingTabla.

\begin{table}[h]
\begin{center}
\begin{tabular}{lcc}
& Accuracy \\ \hline
DMC-Net  & 61.5 \\
DMC-Net w/ Att (CAM)  & 61.4 \\
DMC-Net w/ Att (GBP)  & 61.5 \\
DMC-Net w/ GAN  & \textbf{61.8} \\
DMC-Net w/ GAN w/ Att (CAM)  & 61.5 \\
DMC-Net w/ GAN w/ Att (GBP)  & 61.0
\end{tabular}
\end{center}
\caption{Accuracy on HMDB-51 averaged over 3 splits for the study of the effectiveness of attending to the discriminative regions of optical flow during training our \oursnickname.}
\label{table:att}
\end{table}

\subsubsection{Experimental results}\label{sec:exp_att}

Although it is reasonable and intuitive to attend to the discriminative regions of optical flow during generating DMC, this idea does not offer benefit beyond the GAN method proposed in the main paper.
In Table \ref{table:att}, \textbf{\oursnickname} is only trained with the flow reconstruction loss and the classification loss; \textbf{DMC-Net w/ Att (CAM)} is replacing the flow reconstruction loss in \textbf{\oursnickname} by the above attention-weighted flow reconstruction loss based on the attention map generated by the CAM method;  \textbf{DMC-Net w/ Att (GBP)} is replacing the flow reconstruction loss in \textbf{\oursnickname} by the above attention-weighted flow reconstruction loss based on the attention map generated by the GBP method.
We can see that \textbf{\oursnickname} achieves accuracy comparable with \textbf{DMC-Net w/ Att (GBP)} and \textbf{DMC-Net w/ Att (CAM)}.
But if we equip \textbf{\oursnickname} with the generative adversarial loss, denoted as \textbf{DMC-Net w/ GAN}, the highest accuracy can be achieved.

Furthermore, we explore whether this strategy of making flow reconstruction loss attending to the discriminative regions of flow is complementary to the proposed adversarial loss.
\textbf{DMC-Net w/ GAN w/ Att (CAM)} and \textbf{DMC-Net w/ GAN w/ Att (GBP)} respectively use the attention map generated by the CAM method and the GBP method to weight the flow reconstruction loss in \textbf{DMC-Net w/ GAN}.
But this strategy of attention-weighted flow reconstruction loss hurts the accuracy of \textbf{DMC-Net w/ GAN} and thus is not complementary to the idea of using the adversarial loss.
We believe this is because the DMC generator trained with the original flow reconstruction loss, the classification loss and the GAN loss can already capture sufficient motion information that can be learned from approximating flow and thus explicitly focusing on the discriminative regions of flow does not offer additional benefits.
Consequently, we opt to do not use the attention-weighted flow reconstruction loss in the main paper.

\subsection{More discussions about the speed}



\subsubsection{Speed of DMC-Net (I3D)}

Table 4 in the main paper shows that our DMC-Net implemented using a I3D classifier, denoted as \textbf{DMC-Net (I3D)}, achieves much better accuracy than using a ResNet-18 classifier, denoted as \textbf{DMC-Net (ResNet-18)}.
Note that the speed of \textbf{DMC-Net (I3D)} and the speed of \textbf{DMC-Net (ResNet-18)} are not directly comparable.
ResNet-18 is a frame-level classifier: given an input frame, \textbf{DMC-Net (ResNet-18)} can classify it with the speed at 0.76ms as reported in the Table 3 in the main paper.

However, I3D is a clip-level classifier: during testing, we follow \cite{Kinetics} to feed 250 frames concurrently into a I3D classifier to obtain one action class prediction.
The per-frame inference time of \textbf{DMC-Net (I3D)} is 0.79ms which is slightly slower yet very close to \textbf{DMC-Net (ResNet-18)} (\ie 0.76ms).
But in order to make one action prediction, \textbf{DMC-Net (I3D)} needs to take 0.79x250=197.5ms while \textbf{DMC-Net (ResNet-18)} only takes 0.76ms with the need of only one input frame.

\subsection{Generalize \oursnickname to different compressed video standards}

It is worthwhile pointing out that although we follow CoViAR \cite{wu2018coviar} to specifically use MPEG-4 video \cite{le1991mpeg}, in real applications it would be interesting to develop methods that can handle different video encoding formats.
In the worst case, we can always convert the input video of arbitrary format into MPEG-4 first. On HMDB-51, FFmpeg \cite{FFmpeg} takes 1.13ms in average to convert one frame when processing each video sequentially, still being much faster than extracting flow for the two-stream method.
Table 3 in the main paper shows that the per-frame inference speed of DMC-Net is 0.76ms and that of the two-stream method is more than 75ms.

\subsection{More ablation studies}

In addition to the model analysis in the main paper's Section 4.3, to further validate our design choices, here we present more ablation studies and some strategies that are alternative to the current settings used in the main paper.


\subsubsection{End-to-end learning}\label{sec:e2e}

In the main paper, we train the generator and the classifier in an end-to-end manner with the gradients from the classification loss propagated to not only the classifier but also the generator.
An alternative training strategy is to separate the training of the generator and the training of the classifier.
Concretely, we can first train the generator without the classifier and the classification loss.
Then we fix the generator only for doing inference and then feed the generated DMC into the classification network to train the classifier using the classification loss only.

\begin{figure*}[t]
\vspace{-4em}
\begin{center}
\includegraphics[width=\linewidth]{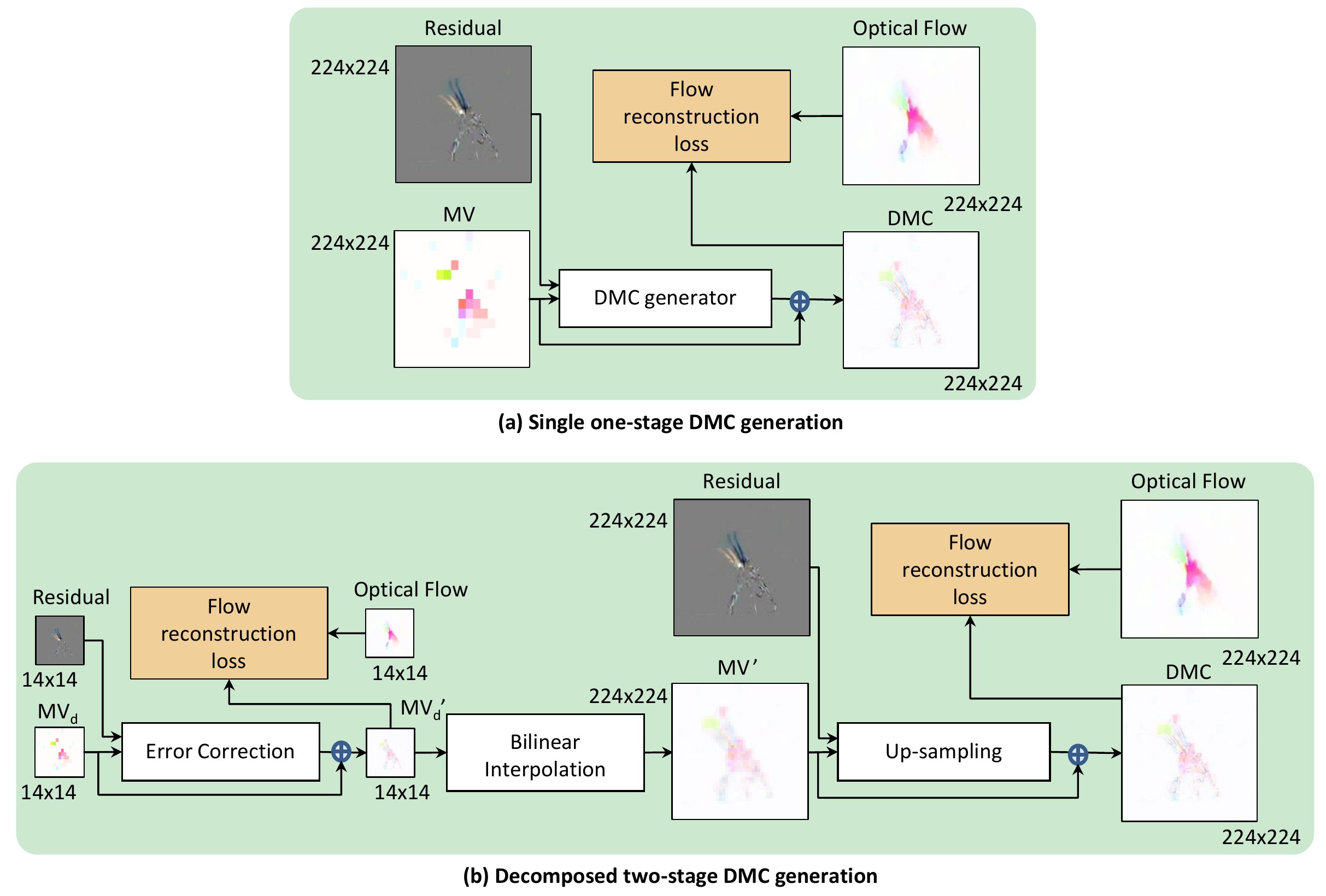}
\end{center}
\caption{Illustrations for (a) the strategy of single one-stage DMC generation used in the main paper and (b) the strategy of decomposed two-stage DMC generation.
 }
\label{fig:framework_twostage}
\end{figure*}

\subsubsection{Decomposed two-stage DMC generation}\label{sec:twostage}

In the main paper, we design a lightweight network to refine Motion Vector (denoted as MV) to generate DMC.
Note that MV has 224x224 spatial size but MV is composed of 16x16 macroblocks in which every pixel has the identical value.
If we downsample MV by a factor of 16 (denoted as MV${_d}$), the same amounts of motion information are still preserved.
Thus the effects of our DMC generator can be considered to be two-fold:
(1) correcting errors and reducing noises in MV and
(2) generating fine details of discriminative motion cue during the process of upsampling MV${_d}$.

Consequently, an alternative way of designing the DMC generator is to first have an error correction network to rectify noises in MV${_d}$ and then have another network to conduct upsampling from 14x14 to 224x224.
As shown in Figure~\ref{fig:framework_twostage} (b), given the stacked residual and MV$_d$ both of size 14x14, we have an error correction network to generate MV$_{d}^{'}$ of size 14x14.
Then the generated MV$_{d}^{'}$ is resized from 14x14 to 224x224 via bilinear interpolation to obtain the MV$^{'}$.
Finally, we feed the stacked residual and MV$^{'}$ into an up-sampling network to generate DMC of more fine motion details.
Note that in Figure~\ref{fig:framework_twostage} (b) we not only measure the flow reconstruction loss between the generated DMC and the corresponding flow but also measure the flow reconstruction loss between the MV$_{d}^{'}$ and the downsampled flow of size 14x14.

\subsection{Smoothing Motion Vector via bilinear interpolation before fed into the DMC generator }\label{sec:upsampling}

As shown in Figure~\ref{fig:framework_twostage} (a), the DMC generator used in the main paper accepts the blocky MV of size 224x224 as input.
Since the optical flow extract by TV-L1 is smooth rather than blocky, smoothing MV before feeding it into the generator can generate DMC of much less blocky artifacts which do not exhibit useful motion information. 
Therefore, instead of directly feeding the blocky MV into the DMC generator, we can make the input MV more smooth by first downsampling MV of size 224x224 to MV${_d}$ of size 14x14 and then resizing MV${_d}$ back to 224x224 via bilinear interpolation.
The rest process follows the main paper.

\subsubsection{Experimental results}

To investigate the effectiveness of the above strategies, we explore different scenarios during the training of \oursnickname using the flow reconstruction loss and the classification loss.
We denote the scenario used in the main paper as \textbf{Ours}, which trains the generator and the classifier in an end-to-end manner, generates the DMC in one single stage, and takes the blocky MV as input for the DMC generator.

First, we compare \textbf{Ours} to \textbf{Ours w/o end-to-end} which follows the above Section \ref{sec:e2e} to separate the training of the DMC generator and the training of the classifier.
Table \ref{table:twostage} confirms the effectiveness of the end-to-end learning strategy and therefore we use it in the main paper.

Second, we compare \textbf{Ours} to \textbf{Ours w/ two-stage} which follows the above Section \ref{sec:twostage} to decompose the DMC generation into a two-stage process.
Table \ref{table:twostage} shows that decomposing the DMC generation into two-stage does not offer benefit in terms of accuracy.
Thus we opt to use the single network in the main paper to generate DMC via jointly correcting errors and generating fine motion details in one single step.

Third, we compare \textbf{Ours} to \textbf{Ours w/ bilinear interp} which follows the above Section \ref{sec:upsampling} to first smooth MV via bilinear interpolation before feeding it into the DMC generator. It turns out that \textbf{Ours} and \textbf{Ours w/ bilinear interp} can generate DMC of comparably good motion cues that lead to similar accuracy.
Therefore in the main paper we directly feed the blocky MV into the DMC generator.

\begin{table}[h]
\begin{center}
\begin{tabular}{ccc}
 & Accuracy \\ \hline
Ours w/o end-to-end & 59.3 \\
Ours w/ two-stage & 60.6 \\
Ours w/ bilinear interp & 61.4 \\
\textbf{Ours} & \textbf{61.5}
\end{tabular}
\end{center}
\caption{Accuracy on HMDB-51 averaged over 3 splits when our \oursnickname is trained with the flow reconstruction loss and the classification loss.}
\label{table:twostage}
\end{table}

\subsection{Results on the full Kinetics dataset}

Due to the extremely long training time on the full Kinectics dataset using one single GPU, we directly adopt the training hyper-parameters used for the Kinetics-n50 subset.
The accuracy of \textbf{CoViAR} is 65.37\%; the accuracy of \textbf{CoViAR + TV-L1 Flow} is 65.43\%; the accuracy of \textbf{DMC-Net (ours)} is 65.42\%.
We can observe that \textbf{DMC-Net (ours)} still improves \textbf{CoViAR} to match the performance of \textbf{CoViAR + TV-L1 Flow} but the performances are very close.
We conjecture this is because when training on such a large-scale dataset, the models for I-frame and Residual have already seen training data of large variance and thus motion information cannot offer significantly complementary cues for distinguishing different action categories.


\clearpage

{\small
\bibliographystyle{ieee}
\bibliography{biblioLong,egbib}
}

\end{document}